# Challenges in Time-Stamp Aware Anomaly Detection in Traffic Videos

Kuldeep Marotirao Biradar, Ayushi Gupta, Murari Mandal, Santosh Kumar Vipparthi
Vision Intelligence Lab, Malaviya National Institute of Technology Jaipur
kuldeepbiradar.cv@gmail.com, ayushigup26@gmail.com, murarimandal.cv@gmail.com,
skvipparthi@mnit.ac.in

## Abstract

*Time-stamp aware anomaly detection in traffic videos is an essential task for the advancement of intelligent transportation system. Anomaly detection in videos is a challenging problem due to sparse occurrence of anomalous events, inconsistent behavior of different type of anomalies and imbalanced available data for normal and abnormal scenarios. In this paper we present a three-stage pipeline to learn the motion patterns in videos to detect visual anomaly. First, the background is estimated from recent history frames to identify the motionless objects. This background image is used to localize the normal/abnormal behavior within the frame. Further, we detect object of interest in the estimated background and categorize it into anomaly based on a time-stamp aware anomaly detection algorithm. We also discuss the challenges faced in improving performance over the unseen test data for traffic anomaly detection. Experiments are conducted over Track 3 of NVIDIA AI city challenge 2019. The results show the effectiveness of the proposed method in detecting time-stamp aware anomalies in traffic/road videos.*

## 1. Introduction

Pervasive use of CCTV cameras in public and private places has laid the foundation for development of various automated systems for intelligent visual monitoring. Numerous tasks such as pedestrian detection, anomaly detection, person re-identification, object tracking, etc. play a significant role in ensuring secure and intelligent transportation. More specifically, automatic detection of anomalous events in road/traffic videos can have multiple applications such as traffic rules violation detection, accidents/suspicious movements analysis, etc. Anomaly/abnormality in videos usually means identification of events that significantly deviate from regular/normal behavior. However, the definition of abnormality may vary according to the context, i.e., time, place and circumstances. For example, driving a car on the road is normal but stalled car on highway is considered to be anomaly. Furthermore, the non-moving cars stationed in parking area does not constitute anomalous behavior. Similarly, the vehicles stopped near traffic lights are normal behavior when it is red but anomaly when it is green. We show samples for different challenging and confusing scenarios in road traffic anomaly detection in Figure 1.

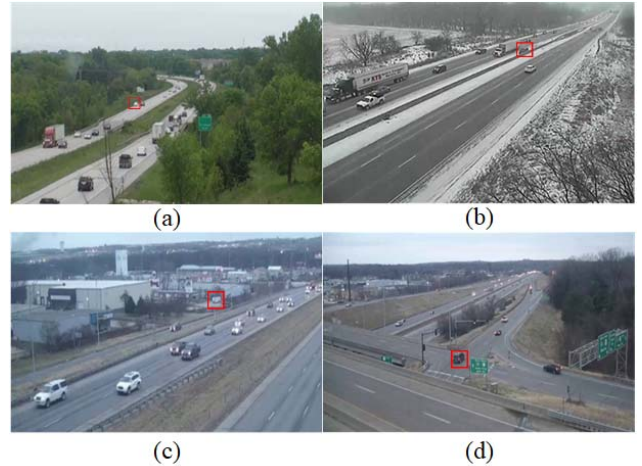

Figure 1. Different vehicle movement/non-movement scenarios in traffic videos. (a), (b) The vehicle stops on the road (anomaly), (c) The vehicle is standing at a parking lot (normal), (d) The vehicle is moving but crossing a red light (anomaly).

Challenges in anomaly detection include appropriate feature extraction, defining normal behaviors, handling imbalanced distribution of normal and abnormal data, addressing the variations in abnormal behavior, sparse occurrence of abnormal events, environmental variations, camera movements, etc. The track 3 of NVIDIA AI city challenge [1-2] presents a carefully designed problem to the researchers to come up with suitable solution and evaluate the same over unseen test videos.

To address the abovementioned challenges for anomaly detection, we propose a deep learning based three-stage pipeline including stages for background estimation, object detection and time-stamp aware anomaly detection. In the first stage, a deep background modelling technique is proposed to estimate the background representation from the recent history. The network learns the object movements in last few frames to differentiate between the static and moving objects. One of the most common anomaly scenarios on roads is when a vehicle stops on the road where it should not (except for when the traffic lights are red). Thus, the estimated background usually consists of the vehicle/vehicles with abnormal behavior. In the second stage, we designed a one-stage object detector to identify the presence of vehicle and traffic lights in the estimated background. In the final stage, we proposed an algorithm to remove temporally inconsistent false positives. The anomaly detection is performed using this time-stamp aware anomaly detector.

We summarize the main contributions of this paper in the following points.
(i) We designed a deep background modelling technique to estimate the background from recent history.
(ii) We designed a one stage object detector to detect the static vehicles and traffic lights from the background image. The idea is to not only detect anomaly in a frame but also localize the anomalous region.
(iii) We design an algorithm to determine the normal/abnormal category for every frame based on the abovementioned two responses. We also present a detailed analysis of the reasons for failure cases of our algorithms.

We evaluate the proposed two-stage model on track-3 test set of the NVIDIA AI city challenge. The experimental result shows that our proposed method can perform reasonably well on the unseen data. We obtain F1-score at 0.3838, RMSE at 93.61 and s3-measure at 0.2641.

## 2. Related Work

Anomaly detection techniques in the literature can be grouped in two categories: traditional and deep learning-based methods. Furthermore, the traditional approaches can be divided in appearance-based and trajectory-based methods. In appearance-based methods, texture features like LBP-TOP [3] is used to extract dynamic encodings. The image is divided into patches where LBP-TOP is applied and dynamic features are extracted from each region. The Bayesian model [4] is applied for the classification of patch based on normal and abnormal events. Similarly, optical flow [5-7], histogram of oriented gradient [8, 9] and histogram of optical flow [9, 10] are also used for anomaly detection.

In trajectory-based methods [11], high level semantic information like speed and direction of moving objects are tracked using selected feature points. Yuan et al. [12] proposed to use 3D DCT model to detect and track pedestrians. Similarly, Lin et al. [13] employed multiple hypothesis tracking algorithm. However, the trajectory of region suffers from detection, segmentation and tracking errors. These errors dramatically increase in crowded or cluttered scenes. In addition, the trajectory of region is computationally expensive in terms of detection and tracking. The appearance-based features are easy to compute and take less time as compared to trajectory-based features.

In recent times, deep learning techniques have shown promising results in various computer vision application including anomaly detection as well. Deep learning models learn optimized set of features through various layers of neural network without requiring any pre-processing. Various applications where deep learning has produced state-of-the-art results include object detection [14], person recognition [15], action recognition [16, 17] and many others. Zhou et al. [18] proposed a 3D convolutional network for anomaly classification. Similarly, Hasan et al. [19] used end to end autoencoders to model temporal regularities in video sequences. In [20], spatiotemporal component is presented where spatial component is used for extracting spatial features and temporal component is used for learning temporal evolution of spatial features. In [21], a framework is proposed where video data is represented by general features. In [22], an unsupervised method is used where appearance, texture and motion features are learned and LSTM is used to detect regularity in the videos. The abnormal events are instances which are different from the modelled regularities. Sabkrou et al. [23] proposed efficient method for detection and localization of anomalies in videos. The authors used transfer learning where the optimized parameters of a supervised CNN are transferred into unsupervised FCN for the detection of anomalies in the scene. Sun et al. [24] proposed a two-stage learning method which utilizes one class learning for detecting abnormalities, the end to end model combines one class SVM with convolution neural network known as deep one class model.

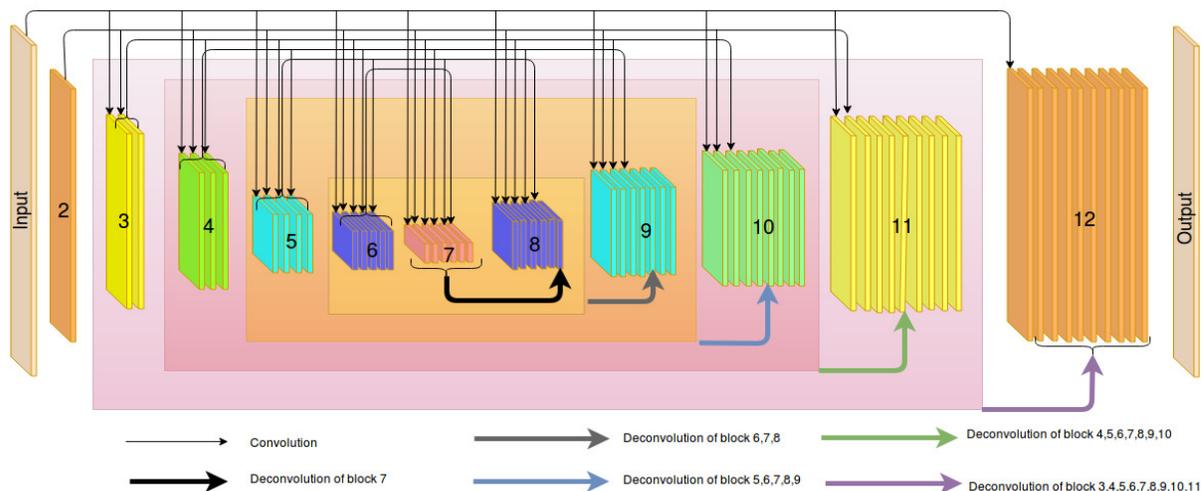

Figure 2. The proposed deep background estimation network.

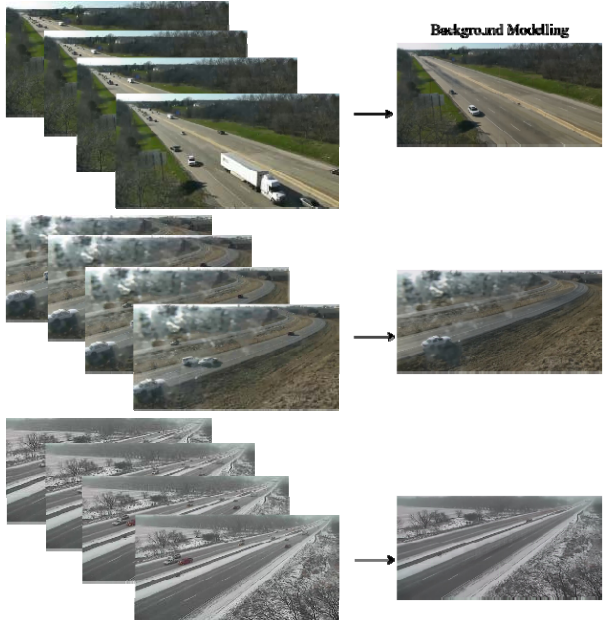

Figure 3. Sample estimated background computed from the proposed deep background estimation network.

In previous NVIDIA AI city challenge, Wei et al. [1] proposed unsupervised anomaly detection method where they used Mixture of Gaussian (MOG) for background modelling. The background estimator removes moving vehicles and keeps the crashed or stopped vehicle as background. Thereafter, the static objects are detected using faster R-CNN for anomaly detection. Similarly, Xu et al. [2] proposed to analyze the vehicle motion pattern in two modes static mode and dynamic mode. In the static mode the vehicle is learned from the background modelling method and extracted using detection procedure to find crashed or stopped vehicle on roads.

## 3. Proposed Method

The detailed description of the proposed two-stage method for time-stamp aware anomaly detection is discussed in the following three subsections: deep background modelling, object detection and the timestamp aware anomaly detector.

### 3.1. Deep Background Modelling

We designed a new CNN based background estimation technique inspired by FlowNet [25] which is used for prediction of optical flow motion vector. Similar to FlowNet, the proposed network is composed of 6 convolutional blocks in encoding stage and 6 deconvolution blocks in the decoding stage. The background estimation network uses 32 kernels of size 3x3 in all the convolutional layers. The network is a two-stage architecture: *diminishing module* and *enhancement module*. The *diminishing module* is composed of various

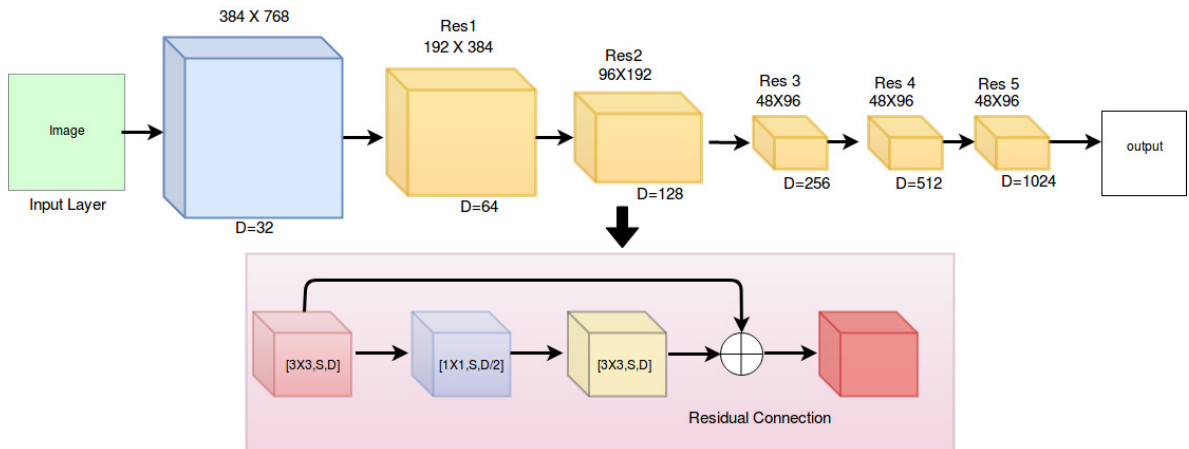

Figure 4. The proposed one-stage object detector for anomalous object localization and classification.
*S=stride, D=feature map depth*

**Algorithm 1** Timestamp aware anomaly detection

**Input:** Vehicle detection response in background image. Let *Vid* contains the set of normal (no detection) and abnormal (one or more detections) labels in a video.
*L*: length (*Vid*)
*N(Win_X): Frequency of normal instances in Win_X*
*A(Win_X): Frequency of abnormal instances in Win_X*
**Initialize:** *Vid$_1$, Vid$_2$, Vid$_3$*
**Output:**
**Step1:**
*Vid$_1$ = Vid*
 **for** *i* in *L*
    **if** (i<5)
       *Win_10 = Vid [0:i]*
    **else**
       *Win_10 = Vid [i-5:i+5]*
    **end**
    **if** *(N(Win_10)>A(Win_10))*
       *Vid$_1$ [i] =normal*
    **end**
 **end**
**Step2:**
*Vid$_2$ = Vid$_1$*
 **for** *i* in *L-20*
    *Win_20 = Vid$_1$ [i:i+20]*
    **if** *(N(Win_20)<5)*
       *Vid$_2$ [i:i+20]=abnormal*
    **elif** *(A(Win_20)<5)*
       *Vid$_2$ [i:i+20]=normal*
    **end**
 **end**
**Step3:**
*Vid$_3$ = Vid$_2$*
 **for** *i* in *L-5*
    *Win_5 = Vid$_2$ [i:i+5]*
    **if** *(N(Win_5)==1)*
       *Vid$_3$ [i:i+5]=abnormal*
    **elif** *(A(Win_5)==1)*
       *Vid$_3$ [i:i+5] = normal*
    **end**
 **end**
**Initial Anomaly Timestamp:**
 **for** *i* in *L*
    **if** (*Vid$_3$ [i] ==abnormal*)
       *Initial anomaly time-stamp = i*3.3 seconds*
       *Break;*
    **end**
 **end**

convolution layers that extracts unrefined to refined features from stacked input images. The feature maps from all the previous layers are stacked at each convolution block (using different strides) while performing feature encoding in the diminishing module. In enhancement module, the detailed information is recuperated through different transpose convolutional layers. To perform refinement, we apply deconvolution to diminishing feature maps from encoding stage and integrate it with corresponding feature maps in the enhancement module. This strategy combines higher level of abstract information from previous layer with information from lower layer feature maps of the network. The proposed deep background estimation model is shown in Figure 2. We also show some sample responses of our background estimator in Figure 3.

### 3.2. Object Detection

After computing the background image for the current frame, we then perform object detection to localize the anomalous region in the image. Since, the object shapes are quite small in most of the videos, we designed a new single-stage object detector inspired by YOLOv2. The proposed object detector is shown in Figure 4. As shown in Figure 4., we used *Res* blocks (residual) at multiple scales to preserve the low-level features present in the shallower layers even while increasing the depth of the network. The proposed network consists of 2 convolutional (conv) layers and 5 residual feature blocks (*Res*). Each *Res* block extracts the salient features by applying two 3x3 and one 1x1 conv operation. These *Res* blocks enhance the capability of the neurons to learn the minute details while maintaining the robustness of the features. All the convolution layers are followed by a batch normalization and leaky ReLu activation layer. We train the object detector for 2 classes: vehicle and traffic lights. If a vehicle is detected, that implies that the current frame consists of anomalous vehicle and thus, the frame is an anomalous frame.

### 3.3. Timestamp aware Anomaly Detection

The object detection response (after background estimation) is used to localize the abnormal region of interest. We then apply the time stamp aware anomaly detection algorithm as given in Algorithm 1. The objective in track3 is to the detect initial time-stamp for anomaly behavior in a video. However, the limitations in 1$^{st}$ and 2$^{nd}$ stage methods sometimes lead to false detection of random noises (signboards, road divider, bushes, etc.) as region of interest in few frames. This results in inconsistent detection of anomaly in a sequence of frames. The proposed Algorithm 1 acts as a postprocessing technique to remove temporally inconsistent false positives to certain degree.

In Algorithm 1, Let's assume the total number of frames in a video *Vid* is *L*. One array is defined for each video which contain possible label (abnormal or normal) for each frame. We explain Algorithm 1 in the following steps.

**Step 1.** Let us consider a middle frame of a temporal window Win_10 as given in Algorithm 1. With reference window, we have validated past and future frames for total number of normal and abnormal frames. If the frequency of normal instances is above 50% then consider the complete window normal, otherwise keep the labels as it is.

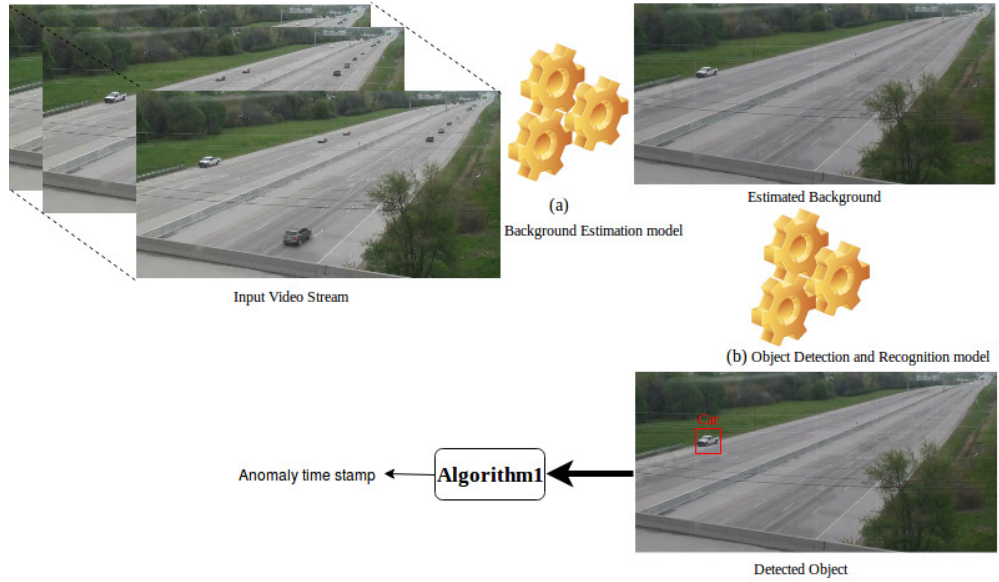

Figure 5. The complete framework for the proposed timestamp aware anomaly detection method.

**Step 2.** We then select a temporal window Win_20 from the responses of **Step 1**. If abnormal instances are above 75% then consider the complete window abnormal and vice a versa.

**Step 3.** Similarly, we select a temporal window Win_5 from the responses of **Step 2**. If abnormal instances are above 80% then consider the complete window abnormal and vice a versa.

Finally, we calculate the initial time of anomaly event in the last stage of Algorithm 1.

## 4. Experimental Results and Discussions

The complete framework of our proposed anomaly detector is shown in Figure 5. The model is trained on train video set of track-3 (NVIDIA AI city challenge). The model is evaluated on test video set of track-3. The anomalies present in track-3 are usually in the form of crashed or stalled vehicles. Each video is recorded for approximately 15 minutes (with 30 fps). These videos consist of diverse backgrounds having rainfall, haze, night time, camera jitter and illumination variations.

**Evaluation Measures.** The results over track-3 test videos are evaluated in terms of F1-score and root mean sum square error (RMSE). The F1-score and RMSE are computed using Eq. (1) and Eq. (2).

$$F1 = 2 \times \frac{(Prec \times Rec)}{(Pre + Rec)} \quad (1)$$

$Pre: Precision, Rec: Recall$

$$RMSE = \sqrt{\sum_{i=1}^{N}(p_i - a_i)^2 / N} \quad (2)$$

where $p_i$ and $a_i$ represent the predicted and actual outcome. $N$ is the total sample size. The ranking in track-3 leaderboard is decided based on the S3-score as computed using Eq. (3)

$$S3 = F1*(1 - NRMSE) \quad (3)$$

where NRMSE denote normalized RMSE.

**Background Modelling.** The training dataset is created by taking every fifth frame from each training video. In this manner 20 frames are selected from 100 frames. These 20 frames are concatenated to form stack of size 384x768x60. Since, the generated stack is too large to be trained on the network directly, therefore, patches of k=128x128 size are extracted from each stack and passed to the network as input layer.

The track-3 train set doesn't provide ground truths for background representation. Thus, for each input stack, temporal median is calculated using 300 frames and used as reference background while training. The mean squared error is generated from the difference between median patch and estimated patch which is back-propagated through the network. The background estimation network uses 32 kernels of size 3x3 in all the convolutional layers.

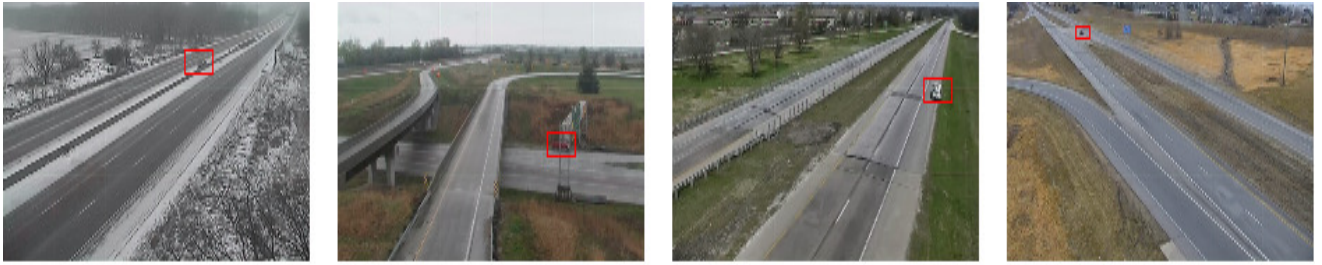

Figure 6. Sample correct results (true positives) achieved by our method. The red boxes represent the anomaly detected when a vehicle is stopped on the road.

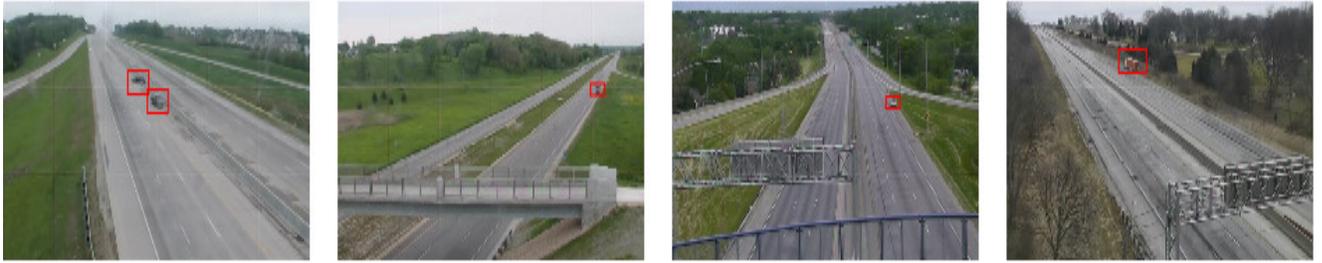

Figure 7. Sample incorrect results (false negatives) of our method. The blue boxes represent the anomaly which should have been detected but our model failed to do so.

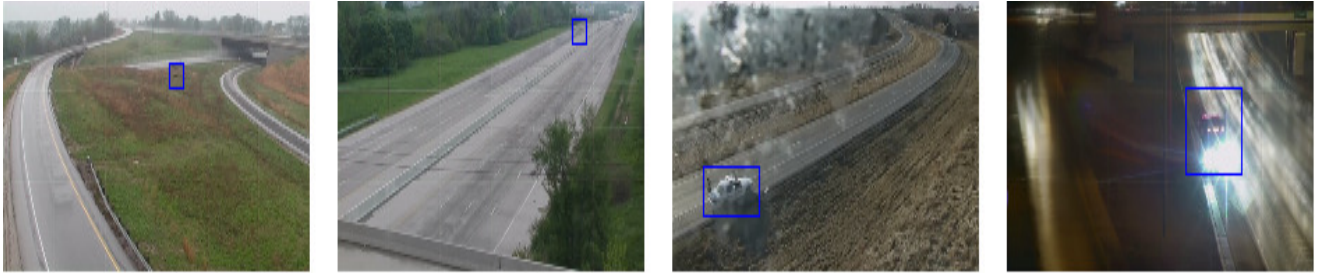

Figure 8. Sample incorrect results (false positives) detected by our method. The red boxes represent the anomaly detected by our model which are just some random noise in the video frame.

The input and output layer shapes are set to 128x128x60 and 128x128x3 respectively. We use Adam optimizer with learning rate $10^{-3}$. At inference time, all the 18 patches generated are concatenated to construct the background frame of size 384x768x3.

**Object Detection.** We prepared bounding box annotations for 1000 samples from training dataset with 2 classes: vehicles and traffic light. The object detector is implemented over the Darknet framework and trained on a Titan Xp GPU. The network is optimized with stochastic gradient descent (SGD) with minibatch size=4. The weight decay and momentum parameters are set to 0.0005 and 0.9 respectively. The inference is performed over the estimated background from background estimation model.

**Qualitative Results.** We show the qualitative results of the proposed method through Figure 6 - Figure 8. In Figure 6, we show the qualitative results for successful anomaly detection by our method. We can see that our method performs well for scenarios where the car is moving on the road and stops after some time. This event is considered as an anomaly. However, due to poor lighting conditions, similarities between foreground and background regions, the object detector failed to detect the anomalous vehicles. This resulted in false negatives for anomaly detection as shown in Figure 7. Similarly, in certain cases, various patches, bushes, etc. are falsely considered as region of interest by the object detector which further increases false positives in the final results.

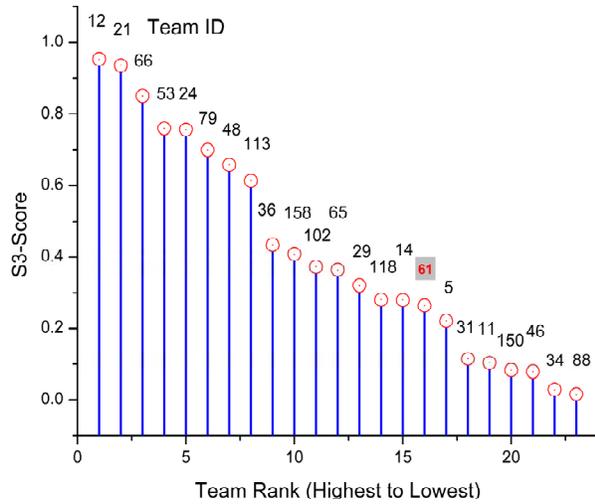

Figure 9. Comparative results of the proposed method and other 23 teams for Track-3 challenge of traffic anomaly detection. (Our Team ID – 61).

**Quantitative Results.** Our method achieved 0.2641 S3-score on track-3 test videos of NVIDIA AI city challenge. It achieved 0.3838 F1-score and 93.61 RMSE respectively. The lowest S3-score is 0.0162. The comparative results are shown in Figure 9.

### 4.1. Analysis of challenges faced while improving performance of the proposed method

**Challenges in background estimation.** There are multiple instances of slow-moving vehicles in certain videos (vid-38 in test set). The vehicle remains in the video for long duration causing misclassification of frames as abnormal. To solve such problems, the model is trained by taking every fifth frame in video. But there is another case of intentional stoppage of videos for some duration which again causes wrong estimation of background. In Figure 10, we show a sample scenario for the case of intentional stoppage.

**Challenges in object detection.** The proposed object detector is able to detect small vehicles but fails to detect large or closely positioned vehicles (video-29 in test set). Some videos are quite blurry which increases the false negative rate of the detector (not even clearly visible through human eyes). Sometimes, the vehicle is detected after certain delays due to which we miss out the initial timestamp of the anomaly. In certain scenarios, the detector could not distinguish between boards, patches, tree and vehicles. So, certain patch is misclassified as anomalous vehicle which causes false positives for anomaly detection. In order to solve this issue, we trained a modified VGG16 classifier to double check the category of the detected objects. However, due to imbalance between number of images in anomalous vehicles and non-anomalous noise data, the classifier failed to achieve much improvements over the object detector. Some sample false detection cases are shown in Figure 7 and Figure 8.

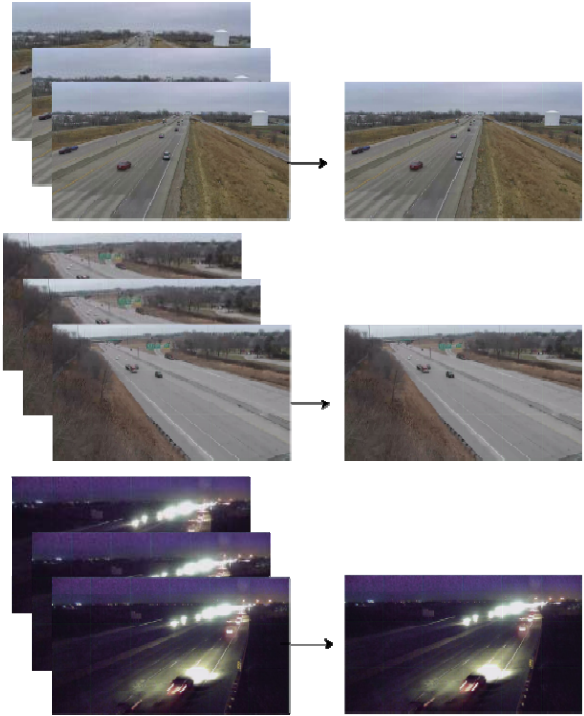

Figure 10. Sample cases for intentional stoppage in videos causing false detections.

## 5. Conclusion

This paper presents a 3-stage pipeline for time-stamp aware anomaly detection in road/traffic videos. A two-stage method was proposed consisting of deep background modelling and one stage object detection. The deep background estimation model learns the object motion patterns based on recent history frames. The proposed background estimation model robustly generates background images in all conditions i.e. camera jitter, rainfall, night vision, etc. The background image is fed through a proposed object detector for anomaly detection. We also present a post-processing technique to remove temporally inconsistent false positives to certain extent. However, in certain scenarios, due to the limitations of background estimator and object detector, we get false positives for patches, signboards, road dividers, etc. Similarly, in few cases, the region of interest is not detected hence F1-score and S-3 are reduced. We proposed an intuitive approach and discussed the challenges to solve the problem of NVIDIA AI city challenge track-3.

## Acknowledgements

This work was supported by the Science and Engineering Research Board (under the Department of Science and Technology, Govt. of India) project #SERB/F/9507/2017. The authors would like to thank the members of Vision Intelligence Lab for their valuable support. We are also thankful to NVIDIA for providing Titan Xp GPU research grant.